# Towards Robust Evaluation: A Comprehensive Taxonomy of Datasets and Metrics for Open Domain Question Answering in the Era of Large Language Models

**Akchay Srivastava, Atif Memon**

Apple Inc., Cupertino, CA 95014, USA

Corresponding author: Akchay Srivastava (akchay_srivastava@apple.com)

This work was supported by Apple Inc.

**ABSTRACT** Open Domain Question Answering (ODQA) within natural language processing involves building systems that answer factual questions using large-scale knowledge corpora. Recent advances stem from the confluence of several factors, such as large-scale training datasets, deep learning techniques, and the rise of large language models. High-quality datasets are used to train models on realistic scenarios and enable the evaluation of the system on potentially unseen data. Standardized metrics facilitate comparisons between different ODQA systems, allowing researchers to objectively track advancements in the field. Our study presents a thorough examination of the current landscape of ODQA benchmarking by reviewing 52 datasets and 20 evaluation techniques across textual and multimodal modalities. We introduce a novel taxonomy for ODQA datasets that incorporates both the modality and difficulty of the question types. Additionally, we present a structured organization of ODQA evaluation metrics along with a critical analysis of their inherent trade-offs. Our study aims to empower researchers by providing a framework for the robust evaluation of modern question-answering systems. We conclude by identifying the current challenges and outlining promising avenues for future research and development.

## I. INTRODUCTION

In Natural Language Processing (NLP), Question Answering (QA) is a core task focused on delivering precise answers to user-posed questions in natural language. This long-standing task dates back to the 1960s [1]. Traditional Machine Reading Comprehension (MRC) systems aim to read and comprehend provided context passages to answer a given question [2]. Conversely, Open Domain Question Answering (ODQA) systems retrieve information from vast unstructured knowledge sources, such as Wikipedia or the Web, to address user queries [3]. This means that the system can answer questions on any topic, such as "Who is the Prime Minister of Japan?" or "What are the different types of clouds?". Compared to MRC, ODQA offers broader applicability and better reflects real-world human behavior. MRC can be considered a stepping stone for achieving ODQA capabilities. Current research in ODQA systems centers around three main approaches, which we now examine.

The first is the retriever-reader [3] approach, which is based on the idea of combining Information Retrieval (IR) and MRC techniques. The retriever component gathers relevant information from external knowledge sources that the reader module leverages to comprehend and formulate an appropriate answer. Notable information retrieval techniques include TF-IDF [3], BM25 [4], and DPR [5]. Transformer-based models have emerged as leading techniques for advanced reading tasks. This includes prominent models such as BERT [6], RoBERTa [7], T5 [8], BART [9], and GPT-3 [10]. Within the retriever-reader ODQA systems, readers can fall into two distinct categories: extractive and generative. Extractive readers directly extract answers from the provided context by identifying the start and end positions of the answer span within the evidence [5], [6]. In contrast to extractive readers, generative readers employ autoregressive token prediction to create fluent answers that extend beyond the given context [8], [11].

The second is the retriever-only approach, which tackles ODQA tasks with a single retriever, eliminating the reading or generating step altogether. A typical category of the retriever-only approach is phrase-based systems [12], [13].

The third is the generator-only approach, which is normally based on single generators, mainly generative Large Language Models (LLMs) such as T5 [8], BART [9], and GPT-3 [10]. Pre-trained on massive Wikipedia corpora, these models encode corpus knowledge within their parameters, thereby enabling direct answer generation without the need for explicit information retrieval. However, evidence has indicated that LLMs can generate hallucinations or other content that contradicts reality [14], [15]. Consequently, ensuring factuality has become a primary concern. The emergence of LLMs has spurred a surge in ODQA research, therefore necessitating a reevaluation of the current landscape.

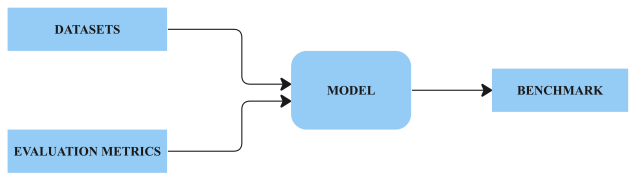

**FIGURE 1.** Role of Datasets and Evaluation Metrics in Benchmarking.

Our study conducts a critical review of datasets and evaluation metrics, two crucial factors influencing the benchmarking of ODQA models, as shown in Fig. 1. Our study did not require the downloading or manipulation of any datasets.

**Related Work:** Several prior reviews have established a strong foundation in this area. For instance, [16] reviewed the construction methodologies used for textual MRC datasets and provided valuable insights regarding the leaderboard details. [17] predominantly explored trends in research architecture for textual ODQA, while also incorporating some examination of the datasets. With a specific focus on the English language only, [18] investigated textual MRC datasets and associated evaluation metrics. Broadening the scope, [19] studied the landscape of datasets and metrics used for evaluating both textual MRC and ODQA tasks. [20] further expanded this reach by encompassing textual and visual QA, with a focus on the datasets only. Building upon recent research in NLP, [21] performed a thorough analysis of the latest textual MRC and QA datasets. While [22] prioritized conversational QA datasets in their investigation, [23] primarily focused on visual QA datasets, offering a more specialized perspective.

Our review distinguishes itself by comprehensively reviewing the datasets and evaluation metrics across all major modalities for both traditional and emerging ODQA tasks. Our review includes the original ODQA datasets, along with hybrid versions that incorporate both external knowledge sources and the provided context. Additionally, we include non-ODQA datasets that have been curated for ODQA purposes. We also include datasets with diverse questions and context topics that are potentially adaptable to ODQA settings through access to external knowledge. We include datasets with answer types including word spans, yes/no responses, and free-form answers. We deliberately exclude datasets with multiple-choice and cloze-style (fill-in-the-missing-word) question formats. We also exclude datasets for closed domain QA, which is limited to specific domains (e.g., medicine, law), and Knowledge Base QA (KBQA), which deals mainly with structured data organized in the form of a knowledge graph.

The main contributions of our review are:

1) **Novel Taxonomy**: We propose a new classification system for textual and multimodal ODQA datasets and their evaluation metrics.

2) **Cross-lingual Dataset Analysis**: To enrich our analysis and provide a more comprehensive understanding of the field, we leverage cross-lingual datasets wherever possible.

3) **Emerging Textual ODQA Tasks**: Our study inaugurates the review of datasets designed for emerging tasks within text-based ODQA. These tasks include counterfactual reasoning, handling ambiguity, time-sensitive information retrieval, and paraphrase comprehension.

4) **Evaluation Metrics for the LLM Era**: Our study paves the way for the review of modern evaluation metrics used for ODQA systems in the context of LLMs. This includes the study of semantic similarity and LLM-based metrics.

5) **Dataset Statistics and Resources**: We provide key statistics for each dataset included in our review, along with pointers to relevant online resources.

**Structure of our Review**: Our review commences with a comprehensive analysis of datasets employed in modern ODQA systems. Section II categorizes these datasets into two distinct classes: textual (Section II-A) and multimodal (Section II-B). A list of links to all publicly available datasets reviewed in our study is provided at the end of Section II. After a comprehensive data analysis, Section III explores the various evaluation metrics used to assess the ODQA system's performance. These metrics are further categorized into two primary approaches: human-based (Section III-A) and automatic (Section III-B) evaluations. In each subsection, a critical discussion is presented that highlights the strengths and limitations of each evaluation technique. Finally, Section IV culminates our work by presenting key observations gleaned from the analysis, pertinent research gaps within the field, and proposing promising avenues for future research endeavors.

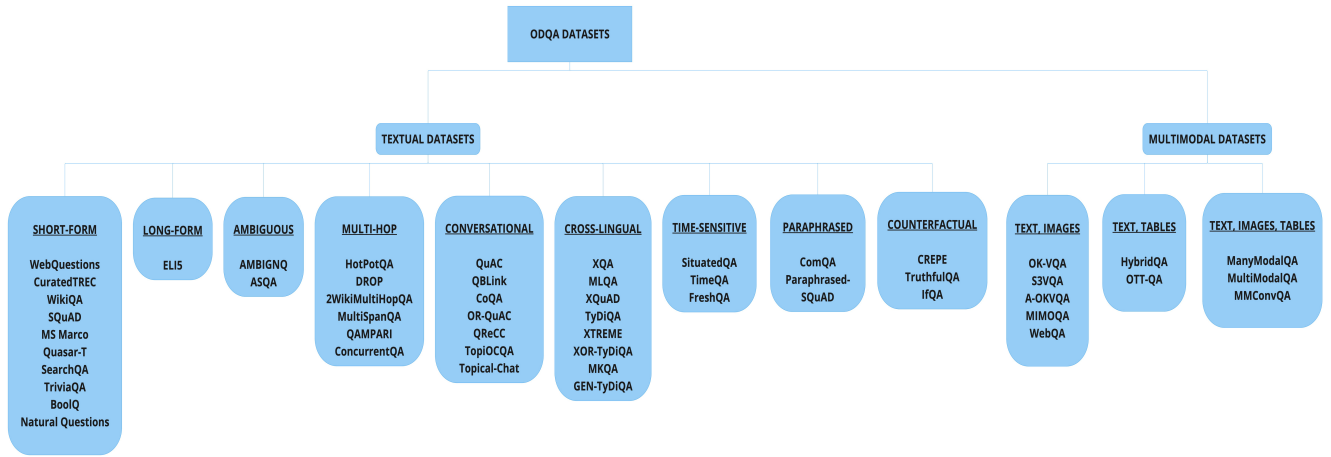

**FIGURE 2.** Modality and Difficulty-Based Taxonomy of Modern Open Domain Question Answering Datasets.

## II. DATASETS

We propose a novel taxonomy for modern ODQA datasets. This classification scheme centers on the modality of both the input data and the knowledge sources that the system leverages. As shown in Fig. 2, our analysis divides the datasets into two fundamental categories: textual and multimodal. To further refine our analysis, we propose a subcategorization of textual datasets based on the question types that they are designed to answer. This approach allows us to explore the spectrum of challenges inherent in these datasets. In the context of multimodal datasets, we propose further sub-classification based on specific modalities integrated within a system.

We present a concise overview of each dataset category, including the publication year, evaluation metrics employed, number of questions, and knowledge source leveraged for answer prediction. This format is consistently applied throughout the study to facilitate comparison across datasets.

### A. TEXTUAL ODQA DATASETS

Textual ODQA systems aim to answer open-ended questions solely based on textual information. In this section, we elaborate on the nine subcategories of the textual datasets within our proposed taxonomy.

#### 1) SHORT-FORM

Short-form questions are unambiguous questions that seek precise factual information and are usually answered in short phrases or sentences. Questions starting with "wh," namely what, when, where, and why, fall under this category. Illustrative examples of short-form questions include "Where was Joe Biden born?" and "What is the capital of California?". These questions that ask for verifiable facts are also known as factoid questions.

##### a) WebQuestions
WebQuestions [24] is a dataset of 6,642 question-answer pairs, where the Google Suggest API was used to generate questions that begin with a "wh" word and contain exactly one named entity. The questions are supposed to be answerable on Freebase, a large knowledge graph. In DrQA [3], WebQuestions was used as an ODQA dataset by changing answer identifiers from Freebase to human-readable entity names and leveraging English Wikipedia as the external knowledge source. The system was evaluated using Exact Match (EM) and F1 scores.

##### b) CuratedTREC
CuratedTREC [25] is a curated version of the TREC QA corpus [26], which is an MRC dataset. TREC stands for the "Text Retrieval Conference," which was started by the U.S. Department of Defense and the National Institute of Standards and Technology in 1992. The dataset consists of 2,180 question-answer pairs from the conference in 1999, 2000, 2001, and 2002. DrQA [3] used CuratedTREC to evaluate open domain questions by leveraging English Wikipedia as the external knowledge source. EM and F1 scores were used for the evaluation.

##### c) WikiQA
WikiQA [27] is a dataset of 3,047 questions sampled from real queries issued on Bing without editorial revisions. The dataset was based on user clicks, and each question was associated with a Wikipedia passage. The initial identification of question-like queries relied on simple heuristics, such as the presence of question words and question marks within the query. It also contained questions that could not be answered using Wikipedia passages. The F1 score was used for the evaluation.

##### d) SQuAD
SQuAD [28], which represents the "Stanford Question Answering Dataset," is an MRC dataset consisting of questions posed by crowd workers on a set of Wikipedia articles. The SQuAD dataset comes in two iterations, launched in 2016 and 2018. SQuAD 1.1 [28] consists of 107,785 question-answer pairs. SQuAD 2.0 [29] adds 53,775 unanswerable questions written adversarially by crowd workers to resemble answerable questions. DrQA [3] used SQuAD for the ODQA task by selecting QA pairs from the development set of SQuAD 1.1, discarding the associated context passages, and instead leveraging the entire English Wikipedia as the external knowledge source. EM and F1 scores were used for the evaluation.

**e) MS Marco**
MS Marco [30] consists of user-submitted questions sampled from the search query logs of Bing or Cortana. The first version was released in 2016 and features 100,000 questions. The most recent version, released in 2018, consists of 1,010,916 questions, along with 8,841,823 passages extracted from 3,563,535 web pages retrieved by Bing. The answers were generated by human editors in their own words instead of selecting a span of text. ROUGE-L [31] and BLEU-1 [32] scores were used to evaluate the system's performance.

**f) Quasar-T**
Quasar-T [33] stands for "Question Answering by Search and Reading (for Trivia)". The Quasar benchmark consists of two datasets. Quasar-S consists of QA pairs based on the software engineering website, Stack Overflow. However, because Quasar-S is a domain-specific dataset, it was excluded from our study. Quasar-T consists of 43,000 open domain trivia questions from multiple internet sources. ClueWeb09 [34], which contains approximately 1B web pages collected between January and February 2009, served as the background corpus for extracting answers. EM and F1 scores were used for the evaluation.

**g) SearchQA**
SearchQA [35] is a dataset that sources questions from the J! Archive (an archive of the American TV quiz show called Jeopardy). The questions were then submitted as queries to the Google search engine. To find potential answers, the system extracted snippets from the top 40 retrieved documents. The answers are short, exact spans of text, typically averaging between 1 and 2 words. The dataset consists of 140,461 QA pairs. Accuracy and F1 scores were used for the evaluation.

**h) TriviaQA**
TriviaQA [36] is a dataset consisting of 95,956 QA pairs written by trivia enthusiasts with independently gathered evidence. Documents for knowledge were collected from the Web and Wikipedia. Trivia-Web used the top 10 Web documents returned by the Bing API as the context for each question. Trivia-Wiki used the content of all Wikipedia pages that contained the entities in the question as the context. EM and F1 scores were used for the evaluation.

**i) BoolQ**
BoolQ [37] is a dataset of 15,942 naturally occurring yes/no questions; that is, the questions are generated in unprompted and unconstrained settings. These questions require a wide range of inference abilities to solve and were gathered from anonymized, aggregated queries issued to the Google search engine. Wikipedia served as the external knowledge source. The system was evaluated using the Accuracy score.

**j) Natural Questions**
Natural Questions [38] consist of 323,045 anonymized and aggregated questions issued to the Google search engine. Human annotators identified different forms of answers within the relevant Wikipedia passages for each question. ORQA [39] used Natural Questions as an ODQA benchmark by only keeping the questions with short answers (answers with 5 tokens or fewer) and using English Wikipedia as the external knowledge source. In practice, this version is most frequently referred to as NQ-Open [39]. EM and F1 scores were used for the evaluation.

Our analysis of short-form datasets reveals a significant trend toward collections specifically curated for ODQA tasks. Several recent studies exemplify this trend by using these curated datasets to benchmark ODQA models. Among these, notable examples are: DPR [5], T5 [8], GPT-3 [10], Fusion-in-Decoder [11], DenSPI [12], ORQA [39], R3 [40], Multi-passage BERT [41], REALM [42], and RAG [43].

Table I summarizes the short-form datasets released between 2013 and 2019. For datasets with multiple versions, the "#Questions" column reflects the count for the most recent one. Short-form datasets vary in size, with question counts ranging from a few thousand to over one million. The most commonly used evaluation metrics are the Exact Match (EM) and F1 scores. However, for datasets containing abstractive answer formats, such as MS Marco, metrics like ROUGE-L and BLEU-1 scores are preferred. Across these datasets, Wikipedia emerges as the dominant knowledge source, although online search engines show increasing prevalence in more recent collections.

TABLE I: SHORT-FORM DATASETS

| Dataset | Year | Metrics | #Questions | Knowledge |
| --- | --- | --- | --- | --- |
| *1. WebQuestions* | 2013 | EM, F1 | 6,642 | Wikipedia |
| *2. CuratedTREC* | 2015 | EM, F1 | 2,180 | Wikipedia |
| *3. WikiQA* | 2015 | F1 | 3,047 | Wikipedia |
| *4. SQuAD 1.1, 2.0* | 2016, 2018 | EM, F1 | 161,650 | Wikipedia |
| *5. MS Marco* | 2016, 2018 | ROUGE-L, BLEU-1 | 1,010,916 | Bing Search |
| *6. Quasar-T* | 2017 | EM, F1 | 43,000 | Web Corpus |
| *7. SearchQA* | 2017 | Accuracy, F1 | 140,461 | Google Search |
| *8. TriviaQA* | 2017 | EM, F1 | 95,956 | Bing Search, Wikipedia |
| *9. BoolQ* | 2019 | Accuracy | 15,942 | Wikipedia |
| *10. Natural Questions* | 2019 | EM, F1 | 323,045 | Wikipedia |

### 2) LONG-FORM

Long-form questions are open-ended questions that cannot be easily answered with a short response or simply by extracting a phrase from a knowledge source. The answers usually consist of multiple sentences or a paragraph. For instance, examples of long-form questions include "How do jellyfish function without brains or nervous systems?" and "How does my car engine work?".

#### a) ELI5

ELI5 [44] is a long-form question answering dataset. The dataset consists of 272,000 open-ended and diverse questions sourced from the Reddit forum "Explain Like I'm Five" (ELI5). Common Crawl's July 2018 web index was used as the external knowledge source. ROUGE-1, ROUGE-2, and ROUGE-L scores were used to evaluate the system's performance.

The long-form datasets are summarized in Table II. It is noteworthy that only one long-form dataset is available for ODQA. Due to the inherent natural language and abstractive nature of long-form answers, various ROUGE-based metrics are used for the evaluation.

TABLE II: LONG-FORM DATASETS

| Dataset | Year | Metrics | #Questions | Knowledge |
|---|---|---|---|---|
| 1. ELI5 | 2019 | ROUGE-1, ROUGE-2, ROUGE-L | 272,000 | Web Corpus |

### 3) AMBIGUOUS

Ambiguous questions are questions that are characterized by multiple possible interpretations or meanings. They hinder the ability of QA systems to arrive at a single, definitive answer because of the lack of sufficient context. Illustrative examples are included in each dataset discussion to aid comprehension.

#### a) AMBIGNQ

AMBIGNQ [45] discussed the challenges of finding clear and unambiguous answers to open-ended factoid questions. A total of 14,042 ambiguous questions were sourced from NQ-Open [39]. The dataset consists of QA pairs, with each answer having a disambiguated rewrite of the original question. The question "Where is the telephone area code 571 located?" is a prime example of an ambiguous question because it is not clear whether the user is asking for counties, cities, or a general area. A disambiguated version of this question provided by AmbigNQ is "*What cities* is the telephone area code 571 located?". Wikipedia served as the external knowledge source. F1 and BLEU scores were used for the evaluation.

#### b) ASQA

ASQA [46], which stands for "Answer Summaries for Questions which are Ambiguous," is a dataset that focuses on ambiguous factoid questions. In total, 6,316 questions were sourced from the AmbigNQ [45] dataset. For example, the question "Who was the ruler of France in 1830?" is ambiguous because France had two leaders in 1830. However, instead of simply disambiguating the question, ASQA leverages disambiguation to identify multiple ground truths and generates detailed answers based on all ground truths. For the above ambiguous question, ASQA would generate "Charles X was King of France from 16 September 1824 until 2 August 1830. France faced urban riots which led to the July Revolution of 1830, which resulted in his abdication and the election of Louis-Philippe I as King of the French. Louis-Philippe was sworn in as King Louis-Philippe I on 9 August 1830." as a detailed answer. Wikipedia was used as the external knowledge source. EM, ROUGE-L, and a disambiguated version of F1 scores were used to evaluate the system's performance.

Table III summarizes the two datasets for evaluating the models on ambiguous open domain questions. AmbigNQ, released in 2020, focuses on predicting multiple possible answers, along with question rewrites that disambiguate the original intent. In contrast, ASQA emphasizes generating detailed, factual summaries that resolve ambiguity. Both datasets leverage Wikipedia as the external knowledge source and employ the F1 score along with an abstractive metric (BLEU or ROUGE-L) for the evaluation. Noticeably, AmbigNQ offers a larger collection of questions.

TABLE III: AMBIGUOUS DATASETS

| Dataset | Year | Metrics | #Questions | Knowledge |
|---|---|---|---|---|
| 1. AMBIGNQ | 2020 | F1, BLEU | 14,042 | Wikipedia |
| 2. ASQA | 2022 | EM, F1, ROUGE-L | 6,316 | Wikipedia |

### 4) MULTI-HOP

Multi-hop questions are questions that cannot be answered directly from a single source of information. These questions necessitate a multi-step reasoning process in which the system iteratively retrieves relevant information from various sources and connects these pieces to arrive at a comprehensive answer. For instance, questions of this nature could include, "Which school that Sir Ernest Rutherford attended has the latest founding date?" and "The actress that had the role of Martha Alston plays what role in Finding Nemo?".

#### a) HotPotQA

HotPotQA [47] is a diverse multi-hop dataset with 112,779 Wikipedia-based QA pairs. Crowd workers were shown multiple supporting Wikipedia documents and were asked explicitly to come up with questions requiring reasoning about all documents. Crowd workers were also asked to supply sentence-level supporting facts that they used to answer the question, allowing for strong supervision, which is also provided as part of the dataset. EM and F1 scores were used as the evaluation metrics.

#### b) DROP
DROP [48], which stands for "Discrete Reasoning Over the Content of Paragraphs," is a crowdsourced dataset covering a variety of categories of Wikipedia pages, particularly those with a high proportion of numbers (sports game summaries and history passages). The dataset consists of 96,567 question-answer pairs. These questions require an ODQA system to be able to resolve references in a question, possibly to multiple input positions, and perform discrete operations over them (such as addition, counting, or sorting). The system was evaluated using the EM and F1 scores.

#### c) 2WikiMultiHopQA
2WikiMultiHopQA [49] is a multi-hop dataset consisting of 192,606 QA pairs. The original questions were sourced from HotPotQA [47] after removing the single-hop and context-dependent multi-hop questions. The final questions were generated programmatically based on predefined templates. Statements from Wikidata (a collaboratively edited open knowledge source) and Wikipedia article summaries that described entities were collectively used as the external knowledge source. EM and F1 scores were used for the evaluation.

#### d) MultiSpanQA
MultiSpanQA [50] consists of questions that require answers to be extracted as multiple discontinuous spans from a text passage. The dataset was created by extracting raw questions and Wikipedia spans from the Natural Questions [38] dataset and then re-annotating them to identify multi-span answers. The dataset consists of over 6,000 multi-span questions in its basic version and expands to 19,608 examples, including unanswerable questions and those with single and multi-span answers. EM and F1 scores were used as the evaluation metrics.

#### e) QAMPARI
QAMPARI [51], which stands for "Questions with Many Answers over Multiple Paragraphs, Indeed," is a crowdsourced dataset of 63,911 questions where the answers are lists of entities spread across many Wikipedia paragraphs. All questions had at least 5 answers, with an average of 13. Examples were generated semi-automatically using the tables from Wikidata and Wikipedia. The questions were then verified to ensure that they could be answered using Wikipedia passages. Finally, the questions were paraphrased in natural language, which makes them suitable for ODQA. The F1 score was used as the evaluation metric.

#### f) ConcurrentQA
ConcurrentQA [52] is a multi-hop crowdsourced dataset of 18,439 questions spanning Wikipedia passages in the public domain and an open-source email collection in the private domain. The format, privacy settings, noise, entity distributions, length of emails, and Wikipedia passages differ in several ways, making this dataset challenging. The system was evaluated using the EM and F1 scores.

The multi-hop datasets released between 2018 and 2023 are summarized in Table IV. The most prevalent evaluation metrics employed are EM and F1 scores. Wikipedia emerges as the most frequent knowledge source across these datasets. However, ConcurrentQA introduces a novel knowledge source by incorporating emails with Wikipedia. 2WikiMultiHopQA stands out with the largest collection of multi-hop questions.

TABLE IV: MULTI-HOP DATASETS

| Dataset | Year | Metrics | #Questions | Knowledge |
|---|---|---|---|---|
| 1. HotPotQA | 2018 | EM, F1 | 112,779 | Wikipedia |
| 2. DROP | 2019 | EM, F1 | 96,567 | Wikipedia |
| 3. 2WikiMultiHopQA | 2020 | EM, F1 | 192,606 | Wikidata, Wikipedia |
| 4. MultiSpanQA | 2022 | EM, F1 | 19,608 | Wikipedia |
| 5. QAMPARI | 2023 | F1 | 63,911 | Wikipedia |
| 6. ConcurrentQA | 2023 | EM, F1 | 18,439 | Emails, Wikipedia |

### 5) CONVERSATIONAL
Conversational questions are questions that emulate natural conversation patterns. In contrast to traditional QA, which is designed to address isolated questions, conversational questions aim to comprehend the overarching context of human dialogue and facilitate a series of interconnected questions.

#### a) QuAC
QuAC [53], which stands for "Question Answering in Context," is a conversational QA dataset in which question-answer pairs are created based on sections of Wikipedia articles. The dataset consists of 98,407 questions and 13,594 conversations. Each conversation involved two crowd workers: the first posed a student who asked a few questions to learn about a hidden passage from Wikipedia, and the second acted as a teacher to answer questions by providing short excerpts from the Wikipedia passage. The questions were more open-ended, unanswerable, or meaningful only within the conversation. The Human Equivalence Score (HEQ) [53] and F1 score were used for the evaluation.

#### b) QBLink
QBLink [54] consists of 56,000 questions for sequential question answering, where the questioner asks multiple related questions about the same concept one-by-one. After each question, the answerer provided an answer before the next question was asked. The questions were fully human-authored, and Wikipedia was used as the external knowledge source. The system was evaluated using the Exact Match score.

### c) CoQA
CoQA [55] is a conversational question answering dataset that consists of passages and human conversations which involves a sequence of question-answer pairs about a passage. The dataset consists of approximately 127,000 question-answer pairs related to about 8,000 conversations. The passages were extracted from documents in seven different domains: child stories, literature, middle and high school English exams, news articles, Wikipedia articles, Reddit articles, and science articles. The F1 score was used for the evaluation of the system's performance. AbgCoQA [56] is an augmented version of the CoQA dataset that contains ambiguous questions.

### d) OR-QuAC
OR-QuAC [57] is a dataset that enhances QuAC [53] by adapting it to an open-retrieval setting, making it more suitable for ODQA. The dataset consists of 40,527 questions related to 5,644 conversations. It is an aggregation of two existing datasets: (1) the QuAC dataset, which offers information-seeking conversations; and (2) the Canard dataset [58], which consists of context-independent rewrites of QuAC questions. The Wikipedia corpus served as the external knowledge source for answering the questions. The HEQ and F1 scores were used for the evaluation.

### e) QReCC
QReCC [59], which stands for "Question Rewriting in Conversational Context," is an open domain conversational QA dataset that consists of 13,700 conversations with 81,000 question-answer pairs. The task in QReCC was to find answers to conversational questions within a collection of 10 million web pages (split into 54 million passages). The answer may span across multiple webpages. EM and F1 scores were used to evaluate the system's performance.

### f) TopiOCQA
TopiOCQA [60] is an open domain conversational dataset with topic switches based on Wikipedia. It consists of 3,920 conversations and 50,574 QA pairs. Conversations start with a real information-seeking question from the Natural Questions [38] dataset, which determines a seed topic (document), and then the questioner may shift to other related topics as the conversation progresses. EM and F1 scores were used for the evaluation.

### g) Topical-Chat
Topical-Chat [61] is an open domain knowledge-grounded conversation dataset without explicit roles for conversation partners and with transitions in conversations. The dataset consists of 10,784 conversations with question-answer pairs embedded in them. External knowledge was collected specifically for 300 popular entities from Wikipedia, Reddit, and the Washington Post. The system was evaluated using the F1 score.

Table V summarizes the conversational datasets released between 2018 and 2023. Common evaluation metrics include EM and F1 scores, with some datasets adopting the novel Human Equivalence Score (HEQ). Wikipedia reigns as the primary knowledge source, while newer datasets like QReCC and Topical-Chat explore websites for this purpose. CoQA stands out as the dataset with the most extensive question collection.

TABLE V: CONVERSATIONAL DATASETS

| Dataset | Year | Metrics | #Questions | Knowledge |
|---|---|---|---|---|
| 1. QuAC | 2018 | HEQ, F1 | 98,407 | Wikipedia |
| 2. QBLink | 2018 | EM | 56,000 | Wikipedia |
| 3. CoQA | 2019 | F1 | 127,000 | Multiple Sources |
| 4. OR-QuAC | 2020 | HEQ, F1 | 40,527 | Wikipedia |
| 5. QReCC | 2020 | EM, F1 | 81,000 | Web Corpus |
| 6. TopiOCQA | 2021 | EM, F1 | 50,574 | Wikipedia |
| 7. Topical-Chat | 2023 | F1 | 10,784 | Wikipedia, Reddit, Washington Post |

### 6) CROSS-LINGUAL
While large and annotated datasets exist for English QA, creating such resources for all languages is practically unfeasible, especially for low-resource ones. Thus, cross-lingual ODQA offers a solution to this challenge. Cross-lingual questions are questions posed in a target language (e.g., Turkish), but the system relies on language-independent features and leverages answers from a source language (e.g., English) with abundant training data.

### a) XQA
XQA [62] is a cross-lingual dataset where the question-answer pairs were sourced from the "Did you know" section of Wikipedia pages. The dataset consists of 90,610 QA pairs in 9 languages: English, French, German, Portuguese, Polish, Chinese, Russian, Ukrainian, and Tamil. EM and F1 scores were used for the evaluation.

### b) MLQA
MLQA [63] consists of QA pairs in 7 languages: English, Arabic, German, Spanish, Hindi, Vietnamese, and Simplified Chinese. It consists of 12,738 extractive QA pairs in English and between 5,029 and 6,006 pairs in each of the other languages. The combined dataset for all languages consists of 46,444 question-answer pairs. Wikipedia article excerpts served as the knowledge source for the system. EM and F1 scores were used for the evaluation.

### c) XQuAD
XQuAD [64] is a dataset that consists of 240 Wikipedia paragraphs and 1,190 question-answer pairs from the development set of SQuAD v1.1 [28], together with their professional translations in 10 languages: Spanish, German,

Greek, Russian, Turkish, Arabic, Vietnamese, Thai, Chinese, and Hindi. EM and F1 scores were used for the evaluation.

#### d) TyDiQA
TyDiQA [65], which stands for "Typologically Diverse Question Answering," is a dataset covering 11 typologically diverse languages (English, Arabic, Bengali, Finnish, Indonesian, Japanese, Kiswahili, Korean, Russian, Telugu, and Thai). The dataset consists of 204,337 question-answer pairs. The questions were generated by annotators by showing them text segments containing the first 100 characters of Wikipedia articles. The annotators were encouraged to ask about anything interesting that came to mind, regardless of whether the questions were unrelated. The F1 score was used as the evaluation metric.

#### e) XTREME
XTREME [66] is a multi-task benchmark for evaluating the cross-lingual generalization capabilities of multi-lingual representations across 40 typologically diverse languages spanning 12 language families. The dataset consists of 142,154 question-answer pairs. These tasks included MLQA [63], XQuAD [64], and TyDiQA-GoldP. TyDiQA-GoldP is the gold passage version of the TyDiQA [65] benchmark, which uses only the gold passage as a context and excludes unanswerable questions. The system was evaluated using the EM and F1 scores.

#### f) XOR-TyDiQA
XOR-TyDiQA [67] consists of 40,000 information-seeking questions across 7 diverse non-English languages (Arabic, Bengali, Finnish, Japanese, Korean, Russian, and Telugu) for which TyDiQA [65] could not find same-language answers. English Wikipedia and the specific language's Wikipedia were used as the external knowledge sources. EM, F1, and BLEU scores were used for the evaluation.

#### g) MKQA
MKQA [68] is a large-scale open domain dataset that consists of 10,000 question-answer pairs across each of the 26 typologically diverse languages (260,000 question-answer pairs in total). English Wikipedia and the specific language's Wikipedia served as the external knowledge sources. EM and F1 scores were used for the evaluation of the system's performance.

#### h) GEN-TyDiQA
GEN-TyDiQA [69] is an open domain generative question answering dataset. The dataset consists of 2,686 human-generated answers for 4,038 questions. It is an extension of the TyDiQA [65] dataset with human-generated, natural-sounding, and complete answers for 5 languages: Arabic, Bengali, English, Japanese, and Russian. Wikipedia was used as the external knowledge source. Accuracy, BLEU, and ROUGE-L scores were used for the evaluation.

Cross-lingual datasets released between 2019 and 2022 are summarized in Table VI. While traditional evaluation metrics, such as EM and F1 scores, remain prevalent, newer datasets have adopted metrics suited to their tasks. For instance, XOR-TyDiQA employs BLEU, and GEN-TyDiQA leverages both BLEU and ROUGE-L to assess the quality of the generated text. Wikipedia serves as the dominant source of knowledge across these datasets. Notably, MKQA boasts the largest question collection, whereas XTREME stands out for its inclusion of the greatest number of typologically diverse languages.

TABLE VI: CROSS-LINGUAL DATASETS

| Dataset | Year | Metrics | #Questions | Knowledge |
|---|---|---|---|---|
| 1. XQA | 2019 | EM, F1 | 90,610 | Wikipedia |
| 2. MLQA | 2020 | EM, F1 | 46,444 | Wikipedia |
| 3. XQuAD | 2020 | EM, F1 | 1,190 | Wikipedia |
| 4. TyDiQA | 2020 | F1 | 204,337 | Wikipedia |
| 5. XTREME | 2020 | EM, F1 | 142,154 | Wikipedia |
| 6. XOR-TyDiQA | 2020 | EM, F1, BLEU | 40,000 | Wikipedia |
| 7. MKQA | 2021 | EM, F1 | 260,000 | Wikipedia |
| 8. GEN-TyDiQA | 2022 | Accuracy, BLEU, ROUGE-L | 4,038 | Wikipedia |

### 7) TIME-SENSITIVE
Time-sensitive questions are those for which the answer can vary depending on the specific time the question is posed. These questions necessitate the system to consider the current time or a designated point in time to deliver an accurate response. Examples of such questions include "How old is Barack Obama?" and "Who won the Oscars for the best actor in 2016?".

#### a) SituatedQA
SituatedQA [70] consists of 11,000 questions that are dependent on time or geography. Time dependent questions were sourced from a variety of existing datasets, such as WebQuestions [24], MS Marco [30], NQ-Open [39], and TyDiQA [65]. Geographically dependent questions were generated separately by modifying existing questions from the NQ-Open dataset based on heuristics. Wikipedia was used as the external knowledge source. EM and F1 scores were used for the evaluation.

#### b) TimeQA
TimeQA [71] is a dataset of 41,172 questions with time-evolving facts. The questions were first identified from Wikidata, and crowd workers then annotated the boundaries of these facts by aligning them with Wikipedia passages. The questions were classified into 'easy' or 'hard' categories based on the difficulty of both temporal understanding and reasoning. The system was evaluated using the EM and F1 scores.

#### c) FreshQA
FreshQA [72] consists of 600 crowdsourced questions that require fast-changing world knowledge, as well as questions with false premises that need to be debunked. The annotators were asked to write questions that involved fresh knowledge and appeared natural. The authors used their dataset to benchmark pre-trained LLMs that do not have access to real-time data or the ability to browse the internet for current information, and they manually evaluated the answers themselves under two different settings.

Table VII summarizes the time-sensitive datasets released between 2021 and 2023. Although traditional evaluation metrics such as EM and F1 scores remain prevalent, newer, potentially smaller datasets like FreshQA have facilitated the use of human evaluation. Wikipedia serves as the primary knowledge source for most of these datasets. Conversely, FreshQA explores a unique approach by leveraging pre-trained LLMs without any external knowledge injection. SituatedQA stands out as the collection that boasts the largest volume of such time-sensitive questions.

TABLE VII: TIME-SENSITIVE DATASETS

| Dataset | Year | Metrics | #Questions | Knowledge |
|---|---|---|---|---|
| 1. SituatedQA | 2021 | EM, F1 | 11,000 | Wikipedia |
| 2. TimeQA | 2021 | EM, F1 | 41,172 | Wikipedia |
| 3. FreshQA | 2023 | Human Rating | 600 | Pre-trained LLMs |

### 8) PARAPHRASED
Paraphrased questions represent formulations of the same fundamental question, using alternative wording or syntactic structures. An example of a paraphrased set of questions is "What causes seasons on Earth?" and "What causes the change of seasons on Earth?".

#### a) ComQA
ComQA [73] consists of 11,214 real-user questions collected from the WikiAnswers community QA website. Questions were grouped into 4,834 paraphrase clusters along with answers through a large-scale crowdsourcing effort, which captures lexical and syntactic variety. Crowd workers were shown pairs of questions from a cluster and were asked to make a binary decision on whether the two questions were paraphrases. Wikipedia served as the external knowledge source. The F1 score was used for the evaluation.

#### b) Paraphrased-SQuAD
Paraphrased-SQuAD [74] consists of 2 sets of paraphrased questions obtained using a paraphrasing model and the paraphrase database PPDB [75]. The first dataset is a collection of 1,062 non-adversarial paraphrases. These questions have been rewritten with minor variations from their original forms. The second dataset is an adversarial paraphrased set that consists of 56 questions paraphrased using context words near a confusing candidate answer. The original questions were sourced from the development set of SQuAD [28], which uses Wikipedia paragraphs as the knowledge source. Semantically dissimilar paraphrases were discarded after the paraphrase generation. EM and F1 scores were used for the evaluation.

The paraphrased datasets are summarized in Table VIII. Traditional evaluation metrics, such as EM and F1 scores, are commonly used. Wikipedia is the primary knowledge source across these datasets. Although published in the same year (2019), these datasets employ distinct paraphrase identification methods. ComQA leverages crowdsourcing to identify paraphrased questions. Conversely, Paraphrased-SQuAD utilizes a paraphrased database and model generation. ComQA boasts a larger collection of questions.

TABLE VIII: PARAPHRASED DATASETS

| Dataset | Year | Metrics | #Questions | Knowledge |
|---|---|---|---|---|
| 1. ComQA | 2019 | F1 | 11,214 | Wikipedia |
| 2. Paraphrased-SQuAD | 2019 | EM, F1 | 1,118 | Wikipedia |

### 9) COUNTERFACTUAL
Counterfactual questions are questions with false presuppositions or false misconceptions. Presuppositions are background beliefs or assumptions that a speaker treats as shared world knowledge. Examples of such questions include "If Los Angeles was on the east coast of the U.S., what would be the time difference between Los Angeles and Paris?" and "Can coughing effectively stop a heart attack?".

#### a) CREPE
CREPE [76] consists of 8,400 Reddit questions, 25% of which contain false presuppositions (especially about unfamiliar topics). Because of the inherent debatability of these false presuppositions, the authors used the most upvoted comments written by community users to annotate the data efficiently. False presuppositions and their corrections were also annotated. Wikipedia was used as the external knowledge source. F1 and BLEU scores were used for the evaluation.

#### b) TruthfulQA
TruthfulQA [77] is a dataset of questions designed to cause imitative falsehoods. To perform well, models must avoid generating false answers learned by imitating human texts. The dataset consists of 817 questions spanning 38 categories, including health, law, finance, and politics. Each question has sets of true and false reference answers and a source that supports the answers. All questions were written by the authors and were designed to be adversarial, i.e., testing for a weakness in the truthfulness of ODQA systems. Pre-trained LLMs were used to generate the answers. Human ratings and LLM-based metrics were used for the evaluation.

### c) IfQA
IfQA [78] consists of questions with counterfactual presuppositions via an "if" clause. The dataset consists of 3,800 questions that were annotated by crowd workers on relevant Wikipedia passages after filtering out those passages that were not related to describing causal events. EM and F1 scores were used for the evaluation.

Counterfactual question datasets, released between 2022 and 2023, are summarized in Table IX. Lexical metrics, such as the F1 score, are commonly used for evaluating factual consistency. CREPE additionally employs the BLEU score for abstractive evaluation, whereas TruthfulQA explores LLM-based evaluation methods. Question sources vary: CREPE leverages online sources such as Reddit, IfQA utilizes Wikipedia passages, and TruthfulQA relies on hand-written questions. CREPE boasts the most extensive question collection. For knowledge sources, Wikipedia is common, with TruthfulQA employing pre-trained LLMs.

TABLE IX: COUNTERFACTUAL DATASETS

| Dataset | Year | Metrics | #Questions | Knowledge |
| --- | --- | --- | --- | --- |
| 1. CREPE | 2022 | F1, BLEU | 8,400 | Wikipedia |
| 2. TruthfulQA | 2022 | Human Rating, LLM-based | 817 | Pre-trained LLMs |
| 3. IfQA | 2023 | EM, F1 | 3,800 | Wikipedia |

### B. MULTIMODAL ODQA DATASETS
In this section, we explore datasets classified according to their utilization of multiple modalities. Multimodal ODQA leverages information from diverse modalities to enhance its comprehension and answer generation capabilities. This allows the system to understand and answer questions in a more comprehensive manner, closer to how humans process information naturally. After text, images and tables are the most commonly used non-textual modalities. The next subsections examine datasets that use textual and image data, followed by datasets that use textual data with tabular data. Finally, the last subsection analyzes datasets that exploit all three modalities: text, image, and table. Examples of multimodal questions include "What is the capital of France and what does it look like?" and "What do these animals usually eat?" (based on Fig. 3(a) and 3(b) as the inputs).

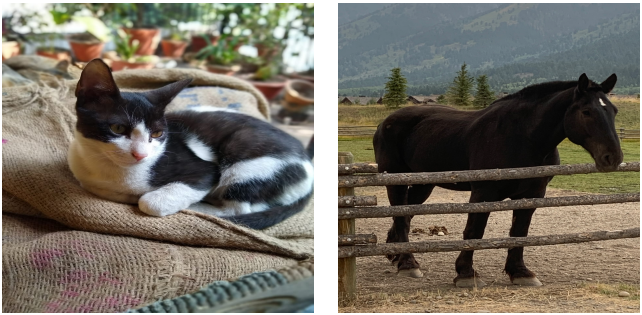

**FIGURE 3(a) and 3(b).** Example of images as inputs for Multimodal ODQA.

### 1) TEXT AND IMAGES
In this subsection, we focus on multimodal datasets that combine text and images.

### a) OK-VQA
OK-VQA [79], which stands for "Outside Knowledge (OK) Visual Question Answering (VQA)," is a dataset consisting of 14,055 open-ended questions where the image content is not sufficient to answer the questions, therefore requiring access to external knowledge. The images were sourced from the COCO [80] image dataset, and passages from Wikipedia articles were provided as external knowledge. The VQA Accuracy [81] score was used for the evaluation.

### b) S3VQA
S3VQA [82] involves the Select, Substitute, and Search (SSS) technique for open domain visual question answering. S3 reaches the end result (i.e., natural language answer) for the VQA-type query by first reformulating the input question (using Select and Substitute) and then retrieving external knowledge source facts (using Search). The dataset is based on the Open Image Collection [83] and consists of 6,765 question-answer pairs. Wikipedia articles were provided as external knowledge. The VQA Accuracy score was used for the evaluation of the system.

### c) MIMOQA
MIMOQA [84] is a dataset in which both input and output are multimodal, incorporating textual and visual elements. Those QA pairs were selected from the MS Marco [30] and Natural Question [38] datasets that were originally extracted from Wikipedia. All images were scraped from the original articles. The dataset consists of 56,693 question-answer pairs. ROUGE and BLEU scores were used as the evaluation metrics for text. Precision@1, 2, and 3 scores were used as the evaluation metrics for images.

### d) A-OKVQA
A-OKVQA [85] is an augmented version of the OK-VQA [79] dataset. It consists of a larger and more diverse set of 24,903 QA pairs, requiring a broad base of both commonsense reasoning and different types of world knowledge to answer. To ease working with unbounded knowledge sources, questions in the training set were paired with rationales that supply facts and snippets of reasoning needed to answer them. The VQA Accuracy score was used as the evaluation metric.

### e) WebQA
WebQA [86] is a multi-hop, multimodal dataset with 46,700 QA pairs. It is structured to mimic human web interaction, which involves asking a question, selecting sources, and generating a fluent language response. The system identifies the relevant information across modalities and combines it with reasoning to answer the query. Wikipedia and Bing image search were used as the external knowledge sources.

The BARTScore [87] for fluency and the F1 score for accuracy were used for the evaluation.

Table X summarizes the multimodal datasets that combine text and images. Although VQA Accuracy remains prevalent, newer datasets explore alternative evaluation metrics. MIMOQA uses abstractive evaluation metrics such as ROUGE and BLEU for text and Precision@k (where k = 1, 2, and 3) for image outputs. WebQA employs the BARTScore metric to assess the semantic similarity of the generated text. Wikipedia continues to be the primary knowledge source. Notably, MIMOQA boasts the most extensive question collection.

TABLE X: TEXT AND IMAGES DATASETS

| Dataset | Year | Metrics | #Questions | Knowledge |
|---|---|---|---|---|
| 1. OK-VQA | 2019 | VQA Accuracy | 14,055 | Wikipedia |
| 2. S3VQA | 2021 | VQA Accuracy | 6,765 | Wikipedia |
| 3. MIMOQA | 2021 | ROUGE, BLEU, Precision@1, 2, 3 | 56,693 | Wikipedia |
| 4. A-OKVQA | 2022 | VQA Accuracy | 24,903 | Wikipedia |
| 5. WebQA | 2022 | F1, BARTScore | 46,700 | Wikipedia, Bing Search |

### 2) TEXT AND TABLES

In this subsection, we focus on multimodal datasets that combine text and tables.

#### a) HybridQA

HybridQA [88] is a QA dataset over both tabular and textual data. Each question is aligned with a Wikipedia table and multiple free-form corpora linked to the entities in the table. The questions were designed to aggregate both tabular and textual information. The dataset consists of 70,000 QA pairs aligned with 13,000 Wikipedia tables. EM and F1 scores were used as the evaluation metrics.

#### b) OTT-QA

OTT-QA [89] is a dataset of 45,814 questions that requires multi-hop reasoning across tabular data and unstructured text. The dataset was built on top of HybridQA [88]. OTT-QA requires a system to retrieve relevant tables and text. EM and F1 scores were used for the evaluation.

In Table XI, we summarize multimodal datasets containing text and tables. Both datasets leverage traditional evaluation metrics such as EM and F1 scores. Notably, Wikipedia serves as the primary knowledge source for these datasets, while HybridQA offers a larger question set.

TABLE XI: TEXT AND TABLES DATASETS

| Dataset | Year | Metrics | #Questions | Knowledge |
|---|---|---|---|---|
| 1. HybridQA | 2020 | EM, F1 | 70,000 | Wikipedia |
| 2. OTTQA | 2021 | EM, F1 | 45,814 | Wikipedia |

### 3) TEXT, IMAGES AND TABLES

In this subsection, we explore multimodal datasets that combine text, images, and tables.

#### a) ManyModalQA

ManyModalQA [90] is a dataset of 10,190 questions for multimodal question answering, which leverages three different modalities: text, images, and tables. The dataset for this challenge was created by scraping Wikipedia and then generating QA pairs through crowdsourcing. A distinctive feature of the dataset is the ambiguity of its questions, which were crafted such that the modality containing the answer is not immediately apparent from the question. The Accuracy score was used for the evaluation.

#### b) MultiModalQA

MultiModalQA [91] is a dataset that requires joint reasoning over text, tables, and images. The questions in the dataset were generated using tables from Wikipedia. The images and text paragraphs of entities that appeared in each table were then attached. The dataset consists of 29,918 QA pairs, 35.7% of which required cross-modal reasoning. EM and F1 scores were used for the evaluation.

#### c) MMConvQA

MMConvQA [92] is a dataset for conversational question answering that uses multiple knowledge sources of different modalities, such as text, tables, and images, through multi-turn conversations. The dataset consists of 1,179 conversations and 5,753 QA pairs. The questions were annotated with natural language answers, corresponding evidence, and decontextualized self-contained questions. EM and F1 scores were used for the evaluation.

Table XII summarizes multimodal datasets containing text, images, and tables. Traditional evaluation metrics like EM and F1 scores remain common. Notably, Wikipedia serves as the primary knowledge source for these datasets. Among them, MultiModalQA boasts the largest collection of question-answer pairs.

TABLE XII: TEXT, IMAGES AND TABLES DATASETS

| Dataset | Year | Metrics | #Questions | Knowledge |
|---|---|---|---|---|
| 1. ManyModalQA | 2020 | Accuracy | 10,190 | Wikipedia |
| 2. MultiModalQA | 2021 | EM, F1 | 29,918 | Wikipedia |
| 3. MMConvQA | 2022 | EM, F1 | 5,753 | Wikipedia |

With this, we conclude Section II, which provided an overview of various datasets for textual and multimodal ODQA. To facilitate further research, we have provided references for accessing the publicly available datasets in Table XIII. The Gen-TyDiQA and MIMOQA datasets are not included in the table because we could not find any public links to them.

TABLE XIII: PUBLIC DATASETS ACCESS LINKS

| Dataset | Link |
| --- | --- |
| 1. WebQuestions | https://github.com/brmson/dataset-factoid-webquestions |
| 2. CuratedTREC | https://github.com/brmson/dataset-factoid-curated |
| 3. WikiQA | https://www.microsoft.com/en-us/download/details.aspx?id=52419 |
| 4. SQuAD | https://rajpurkar.github.io/SQuAD-explorer |
| 5. MS Marco | http://www.msmarco.org |
| 6. Quasar-T | https://github.com/bdhingra/quasar |
| 7. SearchQA | https://github.com/nyu-dl/dl4ir-searchQA |
| 8. TriviaQA | https://nlp.cs.washington.edu/triviaqa/ |
| 9. BoolQ | https://github.com/google-research-datasets/boolean-questions |
| 10. Natural Questions | https://github.com/google-research-datasets/natural-questions |
| 11. ELI5 | https://facebookresearch.github.io/ELI5/ |
| 12. AMBIGNQ | https://nlp.cs.washington.edu/ambigqa/ |
| 13. ASQA | https://github.com/google-research/language/tree/master/language/asqa |
| 14. HotPotQA | https://hotpotqa.github.io/ |
| 15. DROP | https://allenai.org/data/drop |
| 16. 2WikiMultiHopQA | https://github.com/Alab-NII/2wikimultihop |
| 17. MultiSpanQA | https://multi-span.github.io/ |
| 18. QAMPARI | https://samsam3232.github.io/qampari/ |
| 19. ConcurrentQA | https://github.com/facebookresearch/concurrentqa |
| 20. QuAC | https://quac.ai/ |
| 21. QBLink | https://sites.google.com/view/qanta/projects/qblink |
| 22. CoQA | https://stanfordnlp.github.io/coqa/ |
| 23. OR-QuAC | https://github.com/prdwb/orconvqa-release |
| 24. QReCC | https://github.com/apple/ml-qrecc |
| 25. TopiOCQA | https://mcgill-nlp.github.io/topiocqa/ |
| 26. Topical-Chat | https://github.com/alexa/Topical-Chat |
| 27. XQA | https://github.com/thunlp/XQA |
| 28. MLQA | https://github.com/facebookresearch/MLQA |
| 29. XQuAD | https://github.com/google-deepmind/xquad |
| 30. TyDiQA | https://ai.google.com/research/tydiqa |
| 31. XTREME | https://sites.research.google/xtreme |
| 32. XOR-TyDiQA | https://nlp.cs.washington.edu/xorqa/ |
| 33. MKQA | https://github.com/apple/ml-mkqa |
| 34. SituatedQA | https://situatedqa.github.io/index.html |
| 35. TimeQA | https://github.com/wenhuchen/Time-Sensitive-QA |
| 36. FreshQA | https://github.com/freshllms/freshqa |
| 37. ComQA | https://qa.mpi-inf.mpg.de/comqa/ |
| 38. Paraphrased-SQuAD | https://github.com/nusnlp/paraphrasing-squad |
| 39. CREPE | https://github.com/velocityCavalry/CREPE |
| 40. TruthfulQA | https://github.com/sylinrl/TruthfulQA |
| 41. IfQA | https://github.com/wyu97/IfQA |
| 42. OK-VQA | https://okvqa.allenai.org/ |
| 43. S3VQA | https://s3vqa.github.io/ |
| 44. A-OKVQA | https://allenai.org/project/a-okvqa/home |
| 45. WebQA | https://github.com/WebQnA/WebQA |
| 46. HybridQA | https://github.com/wenhuchen/HybridQA |
| 47. OTTQA | https://github.com/wenhuchen/OTT-QA |
| 48. ManyModalQA | https://github.com/hannandarryl/ManyModalQA |
| 49. MultiModalQA | https://github.com/allenai/multimodalqa |
| 50. MMConvQA | https://github.com/liyongqi67/MMCoQA |

## III. EVALUATION

ODQA evaluation can be broadly categorized into two main approaches: Human Evaluation and Automatic Evaluation (Fig. 4). We further decompose Automatic Evaluation into three categories: Lexical (focusing on word-level properties), Semantic (assessing meaning representation), and LLM-based (utilizing large language models for evaluation). To conclude the analysis of each evaluation method, we examine its strengths, limitations, and relevant findings from the existing literature. We will delve into Human Evaluation in Section III-A and Automatic Evaluation in Section III-B.

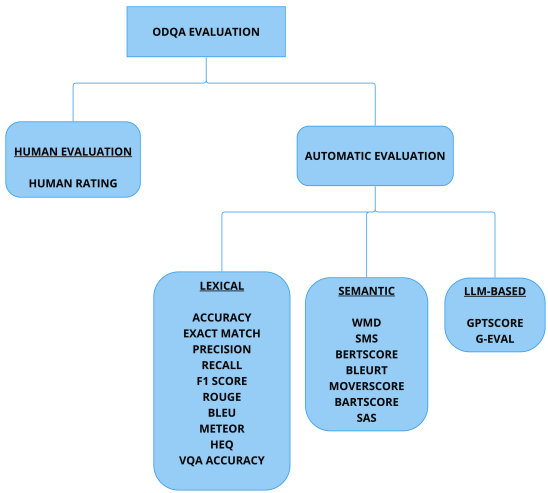

**FIGURE 4.** Taxonomy of Modern ODQA Evaluation Metrics.

### A. HUMAN EVALUATION

Human evaluation provides a holistic assessment of an ODQA system's ability to understand, reason, and generate informative responses that align with the user's intent. Prior studies have relied on human evaluations to assess the quality of QA models [93], [94].

**Human Rating** is the most common form of human evaluation, which refers to the establishment of a well-defined evaluation process involving a recruited group of human raters in the relevant domain. These raters assess the answers based on a pre-determined set of clear and objective criteria. The criteria might include:

- Factual correctness of the answer.
- Relevance to the question and its intent.
- Completeness and informativeness of the answer.
- Clarity and fluency of the language used.
- Originality and insightfulness of the answer.

An interface is set up where raters can see the question, the ODQA system's answer, and provide ratings based on the defined criteria. Some evaluations might involve a binary rating (correct or incorrect) for factual questions, while others might use a Likert scale (1–5) for subjective criteria, such as relevancy, as shown in Fig. 5. Once the evaluations are complete, the data is analyzed to assess the overall performance of the ODQA system. This might involve calculating the average scores across different criteria or identifying areas where the system struggles.

| Strongly disagree | Disagree | Neutral | Agree | Strongly agree |
|---|---|---|---|---|
| 1 | 2 | 3 | 4 | 5 |

**FIGURE 5.** Example of a five-point Likert Scale.

**Strengths**: In the realm of ODQA systems, human evaluation remains the gold standard for assessing system quality. This is due to the inherent ability of human evaluators to assess a system's capability for nuanced language comprehension. Evaluators can effectively judge a system's proficiency in grasping the subtleties of natural language use and crafting responses that are not only factually accurate but also contextually appropriate.

**Limitations**: Inter-rater reliability (IRR) [95], a statistical measure of agreement between multiple raters, is paramount for ensuring the consistency and validity of human evaluations. A low IRR can introduce noise into the data and undermine the generalizability of the findings. Several factors can influence the IRR, such as rater training, subjectivity in ratings, and clarity of the evaluation criteria. Furthermore, human evaluation can be time-consuming and resource-intensive, particularly for large-scale datasets.

**Related Analysis:** The authors of [94] found that the performance of QA systems increased by 23% on average when using human judgments for evaluation.

### B. AUTOMATIC EVALUATION

The automatic evaluation of an ODQA system offers a valuable complement to human assessment. Automatic evaluation refers to the process of using computer programs to assess how well the system performs. This contrasts with human evaluation, in which people judge the quality of the answers. In automatic evaluation, reference answers play a critical role. These high-quality answers represent the expected output for a given question and serve as a benchmark for assessing the performance of the system. We discuss three primary categories of automatic evaluation techniques in the following subsections: Lexical Matching, Semantic Similarity, and LLM-based. Each category provides distinct insights into the performance of the ODQA systems.

#### 1) LEXICAL EVALUATION

Lexical evaluation is used to assess how similar two pieces of text are based on their syntactic vocabulary. This technique focuses on the surface level, looking at characters, words, and n-grams (sequences of words). An n-gram is a way of describing a set of n consecutive words in a sentence.

In the sentence "The sky is blue," we can identify various n-grams, as shown below:

- 1-gram (unigram): "The", "sky", "is", "blue"
- 2-gram (bigram): "The sky", "sky is", "is blue"
- 3-gram (trigram): "The sky is", "sky is blue"
- 4-gram: "The sky is blue"

Note that the words in an n-gram are taken in order, so "the sky blue is" is not a valid 4-gram. Traditional lexical matching methods have been quite popular in ODQA evaluation [3], [11], [39].

#### a) Accuracy
Accuracy reflects the proportion of questions that a system answers correctly, as shown in (1). The definition of correctness can vary depending on the use case and is dependent on the implementation. A question could correspond to one or more correct answers (and could be words, phrases, or sentences).

$$\text{Accuracy} = \frac{\text{Number of Correct Answers}}{\text{Number of Questions}} \quad (1)$$

#### b) Exact Match
Exact Match (EM) refers to the proportion of questions where the system's response aligns precisely with the reference answer, with no word-level discrepancies. Even a small deviation results in a score of 0. Although it is easy to understand and implement, exact matches can be overly strict. Natural language often allows for paraphrasing and slight variations, and a system might be penalized for not capturing the exact wording even if the meaning is conveyed accurately.

#### c) Precision
Precision refers to the proportion of correctly predicted words in the system's response compared to the total number of words in that response, as shown in (2). Precision focuses on accuracy.

$$\text{Precision} = \frac{\text{Number of Correctly Predicted Words}}{\text{Number of Words in the Prediction}} \quad (2)$$

#### d) Recall
Recall refers to the proportion of correctly predicted words in the system's response compared to the total number of words in the reference answer, as shown in (3). Recall focuses on coverage and is also referred to as Sensitivity.

$$\text{Recall} = \frac{\text{Number of Correctly Predicted Words}}{\text{Number of Words in the Reference}} \quad (3)$$

#### e) F1 Score
F1 score is defined as the harmonic mean of precision and recall. It is a widely used metric because it provides a single score that considers both precision and recall, as shown in (4). A high F1 score indicates that the system is both good at finding correct answers (high recall) and avoiding incorrect answers (high precision).

$$F1 = 2 \times \frac{\text{Precision} \times \text{Recall}}{\text{Precision} + \text{Recall}} \quad (4)$$

#### f) ROUGE
ROUGE [31] stands for "Recall-Oriented Understudy for Gisting Evaluation." It is a set of metrics originally proposed to evaluate text generation models (summarization or machine translation). Currently, ROUGE is commonly used for the evaluation of ODQA systems. ROUGE-N is the n-gram recall score between predicted and reference answers. Similarly, ROUGE-L is calculated based on the longest common subsequence between predicted and reference answers. ROUGE-S is a skip-gram concurrence metric that allows one to search for consecutive words from the reference answer that appear in the predicted answer but are separated by one or more other words, thereby adding a degree of leniency. Another variant, ROUGE-W, assesses weighted word overlap, assigning more weight to specific words or phrases. ROUGE emphasizes recall.

#### g) BLEU
BLEU [32], which stands for "Bilingual Evaluation Understudy," is a metric originally proposed for the evaluation of machine translation. The BLEU score is an n-gram precision score between the predicted and reference answers. It incorporates a brevity penalty with an exponential decay that penalizes predicted answers that are too short compared to the reference answers. The brevity penalty addresses the absence of a recall component in the BLEU score.

#### h) METEOR
METEOR [96] stands for "Metric for Evaluation of Translation with Explicit Ordering." Unlike BLEU [32], METEOR considers both precision and recall. This provides a more balanced evaluation. In addition, it also includes features such as synonym matching. The final meteor score combines the F1 score computed from precision and recall with the chunk penalty. Meteor-next [97], a new metric, includes improved text normalization, higher-precision paraphrase matching, and discrimination between content and function words.

#### i) HEQ
HEQ [53] stands for "Human Equivalence Score." It is an evaluation metric that has been used in conversational question answering systems such as QuAC [53]. For systems with multiple valid answers, the F1 score may be misleading. HEQ is an evaluation metric for judging whether the output of the system is as good as that of an ordinary person. Given n references, the average of the maximum F1 computed from each n-1 subset with respect to the held-out reference is considered the F1 score of humans. Following that, let's suppose a QA task contains N questions, and the number of questions for which the token-level F1 performance of the system exceeds or reaches the token-level F1 score of humans is M. The HEQ score can then be computed, as shown in (5).

$$\text{HEQ} = \frac{M}{N} \quad (5)$$

**j) VQA Accuracy**
VQA Accuracy [81] is a commonly used metric in visual question answering that is designed to mitigate inter-rater variability in answer phrasing. It is calculated as the minimum between the ratio of human agreements to a fixed value (typically 3) and 1, as shown in (6). Therefore, this scoring method grants a full point for a question if the system's answer matches that of three or more human annotators. The selection of matching techniques, such as exact matching or another form of lexical matching, depends on the specific application and its requirements.

$$\text{VQA Accuracy} = \min\left(\frac{\text{\#humans that provided the answer}}{3}, 1\right) \quad (6)$$

**Strengths**: Lexical metrics are straightforward to comprehend and implement because they focus on textual features of the language. Furthermore, their well-defined nature allows for easy automation and scalability, making them suitable for large-scale text processing applications.

**Limitations**: Lexical metrics exhibit limitations in their ability to capture semantic nuances. They primarily focus on surface-level text features and struggle to account for variations in text structures, synonyms, and paraphrasing.

**Related Analysis:** Several studies have investigated the applicability of ROUGE and BLEU across various Natural Language Generation (NLG) tasks. [98] demonstrated that, in the context of question generation, lexical metrics exhibit a weak correlation with the concept of answerability. In their work, [99] investigated the limitations of ROUGE and BLEU in evaluating QA systems, particularly for yes-no and entity question types. For such datasets, a single word difference between the generated and reference answer can represent a significant semantic difference (i.e., a correct answer becomes incorrect). Consequently, [99] argued that ROUGE and BLEU scores in these scenarios might not accurately reflect the true quality of the generated answer. Based on their study, [100] indicated that while the F1 score can be a suitable metric for many span-based QA datasets, its performance can be impacted by the specific characteristics of the questions and answers, therefore highlighting the importance of considering these factors when selecting or adapting the evaluation metrics for QA tasks. In addition, [100] advocated the use of the METEOR metric for evaluating generative QA systems. Their findings demonstrate that METEOR achieves the strongest correlation with human judgments of quality.

**2) SEMANTIC EVALUATION**
Semantic evaluation leverages semantic representations to capture the meaning of an answer by focusing on semantic similarity rather than word overlap. This enables semantic metrics to account for variations in phrasing. Evaluation can be seen as a classification task in which the goal is to determine whether a reference answer and a predicted answer are semantically equivalent. These metrics have been instrumental in assessing the effectiveness of various NLG tasks, including machine translation, image captioning [101], dialogue systems [102], and text summarization [87]. For the task of ODQA, the efficacy of semantic metrics has been studied and evaluated by [100], [103], and [104].

**a) Word Mover's Distance**
Word Mover's Distance (WMD) [105] is based on word embeddings that learn semantically meaningful representations for words from local co-occurrences in sentences. The WMD measures the dissimilarity between two text documents as the minimum distance that the embedded words of one document need to "travel" to reach the embedded words of another document. Because WMD uses word embeddings only, it might not fully capture the nuances of context within a sentence (grouping of words).

**b) Sentence Mover's Similarity**
Sentence Mover's Similarity (SMS) [106] is a metric based on sentence embeddings instead of word embeddings, therefore providing the metric access to higher-level representations of text. SMS uses sentence representations that are the averages of their word embeddings.

**c) BERTScore**
BERTScore [101] is an automatic evaluation metric for text generation that first obtains representations for each word in the generated answer and the reference answer by feeding them separately through the BERT [6] (Bidirectional Encoder Representations from Transformers) model and then further proceeds to compute pairwise cosine similarity scores.

**d) BLEURT**
BLEURT [102] is an end-to-end trained metric based on the BERT [6] model. It was designed to model human judgment more effectively. The model is trained on a combination of real and synthetic data. Therefore, its effectiveness is particularly notable in scenarios where training data is scarce or out-of-distribution.

**e) MoverScore**

MoverScore [107] is a metric that takes inspiration from WMD [105] to formulate another optimal matching metric that uses contextualized embeddings to compute the Euclidean distances between words or n-grams. In contrast to BERTScore [101], which allows one-to-one hard matching of words, MoverScore allows many-to-one matching because it uses soft or partial alignments.

**f) BARTScore**
BARTScore [87] is a metric based on the idea that the evaluation of generated text should be considered as a text generation problem, directly evaluating text through the lens of its probability of being generated from or generating other textual inputs and outputs, using pre-trained encoder-decoder models. It can better support the evaluation of generated text from different perspectives (e.g., informativeness, coherence, and factuality) by adjusting the inputs and outputs of the conditional text-generation problem.

**g) Semantic Answer Similarity**
Semantic Answer Similarity (SAS) [108] is a metric that uses a cross-encoder architecture and transformer-based language models that are pre-trained on Semantic Textual Similarity (STS) datasets. The reference and generated answers are separated by a special token to calculate the similarity score.

**Strengths:** Semantic similarity metrics offer an objective approach for quantifying relationships between texts. By capturing the meaning and intent of the language, these metrics enable systems to effectively handle variations in sentence structure, formation, phrasing, and expression.

**Limitations:** Since semantic similarity metrics provide a relative assessment of the similarity between text pieces, they are not capable of determining an absolute measure of truth.

**Related Analysis**: The authors of [100] argued that while semantic similarity metrics perform well for evaluating summarization and translation, that does not necessarily indicate success in QA evaluation since they were shown to exhibit poor correlation with human judgments. Although [104] attempted to improve semantic similarity evaluation by incorporating both question and answer during similarity computation, they also concluded that such metrics remain inadequate for assessing the performance of QA tasks. Furthermore, [104] highlighted the sensitivity of these metrics to the selection of a similarity threshold, which can lead to inconsistencies in the evaluation results. [109] noted that these metrics lack the ability to capture attributability.

### 3) LLM-BASED EVALUATION
Large language models (LLMs) have emerged as a focal point in recent research. This is primarily driven by their proficiency in handling intricate linguistic and reasoning tasks, along with their ability to generate coherent textual outputs. The GPT series, including models such as GPT-3.5 and its chat variant ChatGPT-3.5 (https://chatgpt.com), are prominent examples of this technological advancement. LLMs are evolving beyond answer generation to encompass potential applications for evaluation and judgment tasks. This approach leverages the strengths of LLMs trained on massive amounts of text data to evaluate the outputs generated by other potentially smaller LLMs. The idea is to use LLMs to score the generated answer under the assumption that the LLMs have learned to assign higher probabilities to high-quality and fluent texts. In scenarios where reference answers are unavailable, the internal knowledge of LLMs serves as the primary mechanism for evaluating the plausibility and correctness of the generated answer. However, when reference answers are provided, LLMs can leverage them to verify the accuracy of the generated answer, thereby enhancing the overall reliability, as shown in Fig. 6 and 7. Beyond factual correctness, LLMs can additionally generate scores for qualitative dimensions of the response, such as:

- Relevance: Does the answer directly correspond to the information sought in the question?
- Coherence: Does the answer exhibit a clear and logical structure that facilitates comprehension of the key points?
- Fluency: Does the answer exhibit both natural language fluency and grammatical accuracy?

Furthermore, a growing body of research has demonstrated that LLMs possess a surprising range of emergent capabilities when coupled with appropriate tuning or prompting methodologies. These capabilities include in-context learning [110], chain-of-thought (CoT) reasoning [111], and zero-shot instruction following [112]. This highlights the potential of using such techniques to develop more robust evaluation methods. [103] and [104] studied LLM-based evaluation techniques for the ODQA task.

**a) GPTScore**
GPTScore [113] is a novel evaluation metric that utilizes the emergent abilities (e.g., zero-shot instruction) of generative pre-trained models to score generated texts. GPTScore used GPT3.5 with a model size of 175B, achieving multi-aspect, customized, and training-free evaluation.

**b) G-Eval**
G-Eval [114] is a prompt-based evaluator that uses LLMs with chain-of-thought reasoning [111] to evaluate the quality of generated text in a form-filling paradigm. The task introduction and the evaluation criteria are only fed as prompts, and LLMs are probed to generate a CoT of detailed evaluation steps. The prompt and the generated CoT are then used to evaluate the NLG outputs. The evaluator output is formatted as a form.

You are an impartial judge. You will be given a question which was sent to an Open Domain Question Answering (ODQA) system and you will be given an output that the system produced. You will also be given a reference answer. Your task is to determine a numerical score for the accuracy of the output based on the reference provided. Use the guidelines below to come up with a final score. You must also justify your score.

Question: What is the Statue of Liberty?

Reference: The Statue of Liberty is a colossal neoclassical sculpture on Liberty Island in New York Harbor, within New York City. The copper statue, a gift to the U.S. from the people of France, was designed by French sculptor Frédéric Auguste Bartholdi. The statue was dedicated on October 28, 1886.

Output: A colossal neoclassical sculpture, the Statue of Liberty stands on Liberty Island in New York Harbor. Designed by French sculptor Frédéric Auguste Bartholdi, the copper statue was a gift to the U.S. from the people of France. The statue was dedicated on October 28, 1888.

Scoring Guideline:
Score 1: The output is completely incorrect. It is completely different from or contradicts the provided reference.
Score 2: The output demonstrates some degree of semantic similarity and includes partially correct information. However, the output still has significant discrepancies with the provided reference or inaccuracies.
Score 3: The output addresses a couple of aspects accurately, aligning with the reference. However, there are still omissions or minor inaccuracies.
Score 4: The output is mostly correct. It provides mostly accurate information, but there may be one or more minor omissions or inaccuracies.
Score 5: The output is correct. It demonstrates a high degree of accuracy and semantic similarity to the reference.

**FIGURE 6.** Example of a Prompt sent to the LLM for evaluation.

Score: 4

Justification:

The output shares a high degree of semantic similarity with the reference answer. Here's a breakdown of the accuracy:

- Correct: The output captures the essence of the Statue of Liberty being a colossal neoclassical sculpture on Liberty Island in New York Harbor, a gift from France to the US designed by Bartholdi.
- Minor Inaccuracy: The dedication date mentioned in the output is incorrect (1888 vs. 1886 in the reference).

While the factual inaccuracy with the date brings the score down slightly from a perfect 5, the output conveys most of the important information accurately, making it a mostly correct answer.

**FIGURE 7.** Score and Justification returned by the LLM.

**Strengths**: LLMs demonstrate exceptional promise in evaluating the accuracy of predicted answers, owing to their remarkable capabilities in language processing and reasoning. LLMs can further streamline qualitative assessments by automating time-consuming and resource-intensive tasks that previously relied on human judgment.

**Limitations**: LLMs are inherently susceptible to the way they are prompted. The phrasing, style, and content of a prompt can significantly influence the output of LLMs. [115] noted and examined several potential limitations of the "LLM-as-a-Judge" approach, including position bias, verbosity bias, self-enhancement bias, and limited reasoning ability, as well as mitigation strategies for each one of them. The limitations include the following:

- Positional Bias: LLM judges favor certain positions over others.
- Verbosity Bias: LLM judges favor longer, verbose responses, even if they are not as clear, high-quality, or accurate as shorter alternatives.
- Self-Enhancement Bias: LLM judges may favor the answers generated by themselves.
- Limited capability in grading math and reasoning questions: LLMs are known to have limited math and reasoning capability [116].

**Related Analysis**: In their study, [103] investigated the potential of zero-shot prompting with reference answers as an alternative to human evaluation for assessing ODQA. Their findings suggested that this method holds promise, although it may not be suitable for detecting issues such as unattributability in long-form answers. Similarly, the authors of [104] evaluated the performance of LLM-based evaluations using the GPT-3.5 model for ODQA assessment. Their findings demonstrated that the LLM-based evaluation achieved reasonably good performance across a variety of datasets when compared to other methods. The analysis of LLM-as-a-Judge [115] revealed that strong LLM judges such as GPT-4 [117] can match both controlled and crowdsourced human preferences well, achieving over 80% agreement, the same level of agreement between humans. This study laid the critical groundwork for the development of evaluation frameworks based on LLMs. JudgeLM [118] proposed fine-tuned LLMs as scalable judges to efficiently and effectively evaluate other LLMs in open-ended benchmarks. The JudgeLM models were trained at various scales ranging from 7B to 33B parameters. This fine-tuning process allows the judge models to acquire the ability to distinguish high-quality from low-quality responses within the context of a defined task set. A recent study proposed FACTSCORE [119] to address the challenge of identifying factual errors (hallucinations) generated by LLMs. This novel evaluation method assesses the factual accuracy of the generated answers by decomposing them into individual atomic facts and verifying them against a trustworthy knowledge source. Building on this foundation, SAFE [120] introduced a comprehensive evaluation approach by leveraging an LLM itself to rate the factuality of atomic facts against the Google search engine. While LLM-based judgment remains a nascent field, it presents intriguing possibilities for the future of ODQA evaluation. As LLMs continue their development, they have the potential to emerge as valuable tools for generating comprehensive and multifaceted assessments of ODQA models. However, further research is needed to solidify best practices and understanding in this area.

Our comprehensive analysis demonstrates that each evaluation technique possesses inherent strengths and weaknesses. Consequently, reliance on a single metric for performance assessment is inadequate. A multifaceted assessment approach that incorporates multiple metrics is essential for comprehensively evaluating the efficacy of a QA system.

## IV. CONCLUSION

Our thorough review of 52 datasets and 20 evaluation metrics encompassing a broad spectrum of Open Domain Question Answering tasks reveals several key findings and illuminates promising directions for future research endeavors.

### A. LIMITATIONS OF AUTOMATIC EVALUATION METRICS

The recent success of LLMs is expected to drive advancements in Generative QA systems. However, a key challenge lies in developing automatic evaluation metrics that reflect human judgment of LLM-powered responses. While traditional ODQA relied on lexical metrics, the shift towards semantic and LLM-based evaluations highlights the need for more effective automatic metrics. We anticipate continued research focus in this area. Additionally, LLMs exhibit a concerning tendency to produce hallucinatory outputs: answers that seem plausible yet lack a factual basis. Developing robust and automated methods for detecting hallucinations within LLM responses is an essential avenue for further research.

### B. LACK OF BENCHMARKS IN PUBLIC-PRIVATE DATA SETTINGS

Our study identified a dearth of benchmarks that comprehensively address the challenges of ODQA in hybrid environments, where both public and private data sources are utilized. This paucity presents a significant hurdle for researchers to develop privacy-preserving conversational ODQA agents. As these agents become increasingly prevalent, we expect a future upswing in research towards the creation of novel datasets and evaluation methods that can assess performance within this complex landscape.

### C. SCARCITY OF COMPLEX-QUESTION DATASETS IN TEXTUAL ODQA

While a wealth of robust datasets exist for factoid and short-form textual question answering, our review revealed a critical shortcoming: the lack of datasets that address more intricate and real-world scenarios. These scenarios represent several challenging question types, including long-form, ambiguous, time-sensitive questions, and questions with multiple valid answers.

### D. DATA LIMITATIONS FOR NON-TEXTUAL MODALITIES

The incorporation of non-textual modalities, particularly videos, into ODQA is an emerging area of research and application. While current datasets for video-based question answering are limited in scale and scope, focusing on closed domain tasks, we anticipate an increase in research and the development of datasets that integrate video in a multimodal open domain setting. Our investigation of image and table modalities highlighted a limitation in the availability of datasets that cater to diverse question formats and difficulty levels. Furthermore, our analysis revealed a lack of datasets that provide support for languages beyond English for non-textual modalities. We anticipate that as this field progresses, there will be a shift towards the development of more comprehensive datasets that bridge these current gaps.

### E. FUTURE DIRECTIONS

Based on the trends observed in this field, we posit that research on multimodal datasets and advanced evaluation metrics will continue to be a focal point. These advancements will be instrumental in paving the way for artificial general intelligence (AGI) by enabling the creation of systems that can effectively process and answer complex questions across diverse modalities, akin to human intelligence.

In this study, we presented a comprehensive review of datasets and evaluation metrics encompassing various modalities for ODQA. We hope that this analysis will serve as a valuable resource for researchers, aiding in the development of novel QA systems and promoting the advancement of robust evaluation methodologies.

## ACKNOWLEDGMENT
We are grateful to Ceren Gerzic, Rajani Samidi, and Stephen Pulman for their invaluable feedback on this paper. We would also like to thank Jessica Wang, Kevin Hsu, and Mohamed Soliman for their support.